\documentclass[preprint,12pt]{elsarticle}

\usepackage{algorithm}
\usepackage{algpseudocode}
\usepackage{xcolor,colortbl}
\usepackage{amssymb}
\usepackage{comment}
\usepackage{hyperref}

\begin{document}

\begin{frontmatter}



\title{Characterizing the contribution of dependent features in XAI methods}


\author[1,2]{Ahmed Salih}
\author[3]{Ilaria Boscolo Galazzo}
\author[1,4]{Zahra Raisi-Estabragh}
\author[1,4,5,6]{Steffen E. Petersen}
\author[3]{Gloria Menegaz *}
\author[7,8]{Petia Radeva *}

\affiliation[1]{organization={William Harvey Research Institute, NIHR Barts Biomedical Research Centre, Queen Mary University of London},
            addressline={Charterhouse Square}, 
            city={London},
            postcode={EC1M 6BQ}, 
            state={London},
            country={United Kingdom}}

\affiliation[2]{organization={Department of Computer Science, Faculty of Science, University of Zakho},
            addressline={Zakho Way Road}, 
            city={Duhok},
            postcode={42002}, 
            state={Kurdistan region},
            country={Iraq}}

\affiliation[3]{organization={Department of Engineering for Innovation Medicine, University of Verona},
            addressline={Via S. Francesco, 22}, 
            city={Verona},
            postcode={37129}, 
            state={Verona},
            country={Italy}}

\affiliation[4]{organization={Barts Heart Centre, St Bartholomew’s Hospital, Barts Health NHS Trust},
            addressline={West Smithfield}, 
            city={London},
            postcode={EC1A 7BE}, 
            state={London},
            country={United Kingdom}}

\affiliation[5]{organization={Health Data Research},
            city={London},
            state={London},
            country={United Kingdom}}

\affiliation[6]{organization={Alan Turing Institute},
            city={London},
            state={London},
            country={United Kingdom}}

\affiliation[7]{organization={de Matemàtiques i Informàtica, Universitat de Barcelona},
            addressline={Gran Via de les Corts Catalanes}, 
            city={Barcelona},
            postcode={08007}, 
            state={Barcelona},
            country={Spain}}

\affiliation[8]{organization={Computer Vision Center, Cerdanyola del Vallés},
            city={Barcelona},
            state={Barcelona},
            country={Spain}}

\affiliation[]{organization={*: Gloria and Petia sharing PI}}

\begin{abstract}
Explainable Artificial Intelligence (XAI) provides  tools to help understanding how the machine learning models work and reach a specific outcome. It helps to increase the interpretability of models and makes the models more trustworthy and transparent. In this context, many XAI methods  were proposed being  SHAP and LIME the most popular. However, the proposed methods assume that used predictors in the machine learning models are independent which in general is not necessarily true. Such assumption casts shadows on the robustness of the XAI outcomes such as the list of informative predictors. Here, we propose a simple, yet useful proxy that modifies the outcome of any XAI feature ranking method allowing to account for the dependency among the predictors. The proposed approach has the advantage of being model-agnostic as well as simple to calculate the impact of each predictor in the model in presence of collinearity. 
\end{abstract}




\end{frontmatter}


\section{Introduction}
Explainable artificial intelligence (XAI) emerged to uncover the mystery around machine learning models, precisely the advanced complicated models including deep neural network and conventional neural networks. XAI helps to understand how these models reached a specific outcome, how each feature or region within an image contributes to the model output and to what extent the model is certain to the reached decision~\cite{gunning2019xai}. Such concepts and principles are indispensable to the end user as they increase the trust and transparency in the model outcome~\cite{szabo2022clinician}. There have been many XAI methods proposed to break down the machine learning models into more explainable models.\\ 
SHAP (SHapley Additive exPlanations) is a XAI model-agnostic method that was proposed based on game theory~\cite{lundberg2017unified}. It considers each feature in the model as a player and the outcome is the payout. Then, it calculates the marginal effect of each feature to the model outcome by calculating a SHAP score based on the game theory/coalition. Local Interpretable Model-Agnostic Explanations (LIME)~\cite{ribeiro2016should} is another XAI method that explains the model locally for a single instance. It shows how each feature contributes to the model outcome for a specific subject and to what extent the model is certain. Partial Dependence Plot is another XAI approach that shows the marginal effect of a feature on the model outcome~\cite{friedman2001greedy}. It shows whether the relationship is monotonic, linear or more complex. Classical approaches were also considered to reveal the effect of the features to the output by reporting the weights (coefficient) of the features in the model~\cite{belle2021principles}.\\
One of the main issues that XAI methods is facing, is the dependency among the used predictors. This has considerable impact on the XAI outcome when providing the most informative predictors in the model. For instance, when SHAP calculates the score (impact) for each predictor, it assumes the features are independent~\cite{aas2021explaining}. In this way, if two features are correlated, one of them will have a high score while the other will get a low score, because it does not improve the model performance due to the dependency with the other predictor. Similarly, LIME and classical machine learning models use the weight of each feature in the model to reveal its impact. The weight represents the dependency of a feature on the outcome while holding all the other predictors in the model constant. Such approach is not precise as group of predictors (collinear) in the model might change simultaneously. In real life applications, it is very rare that the used predictors in the model are independent. Accordingly, the outcome of XAI methods such as the list of informative predictors should consider the dependency to provide a more robust list. Here, we propose a new proxy/method that takes the outcome of any feature ranking XAI method and modifies it according to the collinearity among the predictors to generate a new ordered list of most informative features in the model.
\section{State of art}
Dependency among the predictors and its impact have been recently faced in several works. Kjersti Aas et al~\cite{aas2021explaining} proposed four approaches to handle the collinearity issue in Kernel SHAP~\cite{lundberg2017unified}. The main contribution of the Kernel SHAP is that the conditional distribution of subsets of features $\varsigma$ is conditional on the rest of features in \textit{S}, where \textit{S} is the whole set of features. In the original form of Kernel SHAP, they implicitly assume that the features are independent, so that it is possible to replace the conditional distribution by the marginal distribution of the predictors in $\varsigma$. Kjersti Aas et al~\cite{aas2021explaining} modified the original Kernel SHAP by estimating the distribution of all features using four approaches based on the features distribution. The considered cases included the Gaussian distribution, Gaussian copula distribution, Empirical conditional distribution and the combination of the empirical approach and/or the Gaussian or the Gaussian copula. Despite their method addresses one of the significant issue, it is expensive in terms of the computation, which 
affects the exploitability of such methods. In addition, the method was proposed to explain an individual prediction (local explanation) rather than a global explanation, where the outcome shows the overall contribution of each feature for all instances in the model. Moreover, the chosen distribution might affect the method outcome which means the end user should be aware of the distribution of the used data in order to chose the right distribution, which in general is not true.\\
Another approach was proposed to handle the dependency among the predictors named Shapley cohort refinement~\cite{mase2019explaining}. Their method consists in creating a similarity cohort that combines instances (data points) close to the target instance (the one that needs to be explained) and then applying the Shapley method to the new cohort. This method classifies the subjects as either similar to the target subject or dissimilar. The similarity is measured by comparing the values of predictors for each subject with the target subject. If the predictors are binary, then it is easy to compare and identify similar subjects. In case the predictors are continues variables, they used a distance based method to identify the subjects that are similar to the target subject. The chosen similarity measure will have significant impact on identifying the importance of each predictor in the model. The method has many advantages including using real data (predictors) and gives similar score to two predictors if they are highly correlated. However, the method heavily depends on the chosen similarity measure which impacts the outcome of the method.
\section{Materials and Methods}
\subsection{Proposed approach}
One of the main significant aspects of a XAI method or proxy to evaluate any XAI method is the simplicity. It is vital for the XAI or the evaluation criterion to be simple as most of the end users are not experts in XAI methods but are rather from different domains or applications. Any end user from any domain needs to understand how the XAI works to increase the trust with the XAI outcome, Indeed, this was the purpose when XAI was proposed: to increase the trust with the machine learning models. Here, we aim at presenting a simple approach for modifying the outcome of any XAI method in a simple way allowing to handle the collinearity across features and yet being easy to understand.\\
The proposed approach is an extension to the previously proposed proxy to assess the stability of XAI methods~\cite{salih2022investigating}. In the previous work, a new criterion was proposed named Normalized Movement Rate (NMR) which measures how the list of the informative predictors generated by any XAI method will be affected when the top predictor is iteratively removed from the model. The NMR value ranges between 0 and 1, where 0 indicates that the informative list is robust, while a value close to one indicates that the list is not stable. This is caused by the dependency among the predictors, representing the main weakness of the method. The NMR value shows that after applying Principal Component Analysis (PCA), the outcome (list of top predictors) generated from XAI methods is more stable and robust. In that work, they only presented a new criterion to diagnose the instability of the XAI outcome due to collinearity among the predictors. In this paper, we move a step forward and  present a novel way to modify the XAI outcome allowing to overcome such a limitation and take into account the dependency among the predictors.\\
Similar to~\cite{salih2022investigating}, here we iteratively remove the top predictors and re-train the model before applying the XAI method on the test data to generate the list of informative predictors. Here we introduce a new index called Modified Informative Position (MIP). We keep removing, re-training and applying the chose XAI method till we have 2 predictors in the model. For each iteration (removing the top predictor), we normalize by dividing the index (representing its significance) of each predictor by the number of predictors in the current model. Finally, we sum up the division of the indexes of each predictor on the number of predictors in each iteration to calculate a score for each predictor. Finally, we sort the predictors based on the score in ascending order. The predictors with smaller scores are more informative as they reached the top list faster than others, while those with higher score are less informative as they were present in the model in most of the iterations. Equation~\ref{MIP} shows how MIP is calcuated for each predictor in the model.
\begin{equation}\label{MIP}
\mathrm{MIP} = \sum_{j=1}^{NF-2}x_{p,j}
\end{equation}
where $j$ is the number of iterations, $NF$ is the number of predictors, $x_{p}$ is the result of division the index $i$ by the number of predictors $p$ in each iteration.\\
Algorithm~\ref{alg:cap} shows the pseudo-code of the proposed method. We also calculate the Standard Deviation (SD) using the score of features. The higher the SD, the more robust the list is and the NMR is smaller. In the best scenario, SD will be the maximum and NMR will be the minimum (close to zero). In addition, if the features in the model are independent (usually not), the outcome of XAI (e.g. SHAP) will be very similar to that of the proposed method. 
\begin{algorithm}[H]
\small
\caption{Pseudocone for calculating the NMR and MIP values. The blue fonts identify the new steps added to the previous method.}\label{alg:cap}
\begin{algorithmic}
\Ensure NF = total number of features number of features
\Require Train the model
\Require Test the model
\Require Apply SHAP and rank the feature importance in descending order
\While{$NF \neq 2$}
\State \textcolor{blue}{Divide the position (index) of each predictor by the number of features: x\textsubscript{p} = i\textsubscript{p}/NF, where i is the index of the predictor p}
\State Remove the most informative predictor
\State Apply SHAP and rank the feature importance in descending order
\State Check how many predictors changed their positions compared to the list at the previous step
\State Calculate the movements of the predictors based on the index
changes and sum them up ($M$)
\State Calculate the maximum possible movements: $MPM = 2\sum_{i=1}^{i+2}(NF - i)$ for $i<NF$
\State Calculate the movement rate: $MR = M/MPM$
\EndWhile
\State $NMR = \sum(MR) / NF_{total}-2$
\Require \textcolor{blue}{Calculate the modified informative position (MIP)}
\For{\texttt{\textcolor{blue}{p in P, where P is the set of predictors}}}
\State \textcolor{blue}{$MIP = \sum_{j=1}^{NF-2}x_{p,j}$ if $p$ exists and $j$ is the number of iterations}
\EndFor
\end{algorithmic}
\end{algorithm}
\subsection{Data}
To illustrate the proposed method, we used data from United Kingdom Biobank~(application number 2964). The used data consisted of nine cardiac phenotypes extracted from Cardiac magnetic resonance imaging including Left Ventricular (LV) End-Diastolic and End-Systolic Volumes (LVEDV and LVESV, respectively), LV Stroke Volume (LVSV), LV Ejection Fraction (LVEF), LV Mass (LVM), Right Ventricular (RV) End-Diastolic and End-Systolic Volumes (RVEDV and RVESV, respectively), RV Stroke Volume (RVSV) and RV Ejection Fraction (RVEF) for 2,000 subjects (1,032 females). The nine phenotypes were used as predictors to classify the subjects into male and female. Those features were chosen on purpose as sex causes variations in their values~\cite{petersen2017reference}.
\subsection{Implementation}
In~\cite{salih2022investigating}, it has been shown that the XAI outcome is model-dependent suggesting that each model might deal differently with the collinearity among the predictors such that in return the XAI method will generate different rankings. Accordingly, we chose a simple task with binary outcome and assessed the stability of the feature ranking across five classifiers. The following classifiers were considered: Support Vector Classifier (SVC), Light Gradient Boosting Machine (LGBM), Decision Tree (DT), Random Forest (RF) and Logistic Regression (LR). The data was divided into training, to train the model, and unseen data, to test (20\%) the model. To chose the optimal parameters for each of the five models, we applied hyperparameter tuning on the training data, using 10 folds cross-validation and set of parameters (e.g. max\_depth in DT, C in SVC, etc) for each model. Accuracy was used as criterion to select the optimal parameters for each model. Then, the model with the optimal parameters was used to test the model on the test data. Finally, SHAP was applied as XAI method to generate the list of informative predictors to reveal which predictors have more impact on the model outcome.\\
\subsection{Validation}
To validate the proposed method, we applied PCA to generate independent components from the original predictors. The number of components was chosen automatically to explain the 95\% of the variance which resulted in 4 principal components (PCs). Then, these components were used to feed the five models as for the original features. Finally, SHAP was applied to order the PCs based on their contributions.
\subsection{Biological plausibility assessment}
To evaluate the proposed method clinically, we used the absolute sex difference value of the nine measures reported in table 1~\cite{st2022sex}. The absolute value of the sex difference was used to order the features in descending order, indicating the higher the score the more the feature is significant to distinguish between male and female. We called this list benchmark to distinguish it from the SHAP list and the list of the proposed method. Thereafter, we performed Kendall's tau\_b rank correlation and Pearson correlation between the order of the nine features in the benchmark, SHAP and the proposed method to reveal whether the SHAP list or the proposed method list is more inline with the benchmark order. 
\section{Results}
\begin{figure}[H]
\centering
\includegraphics[width=\textwidth, height= 5 cm]{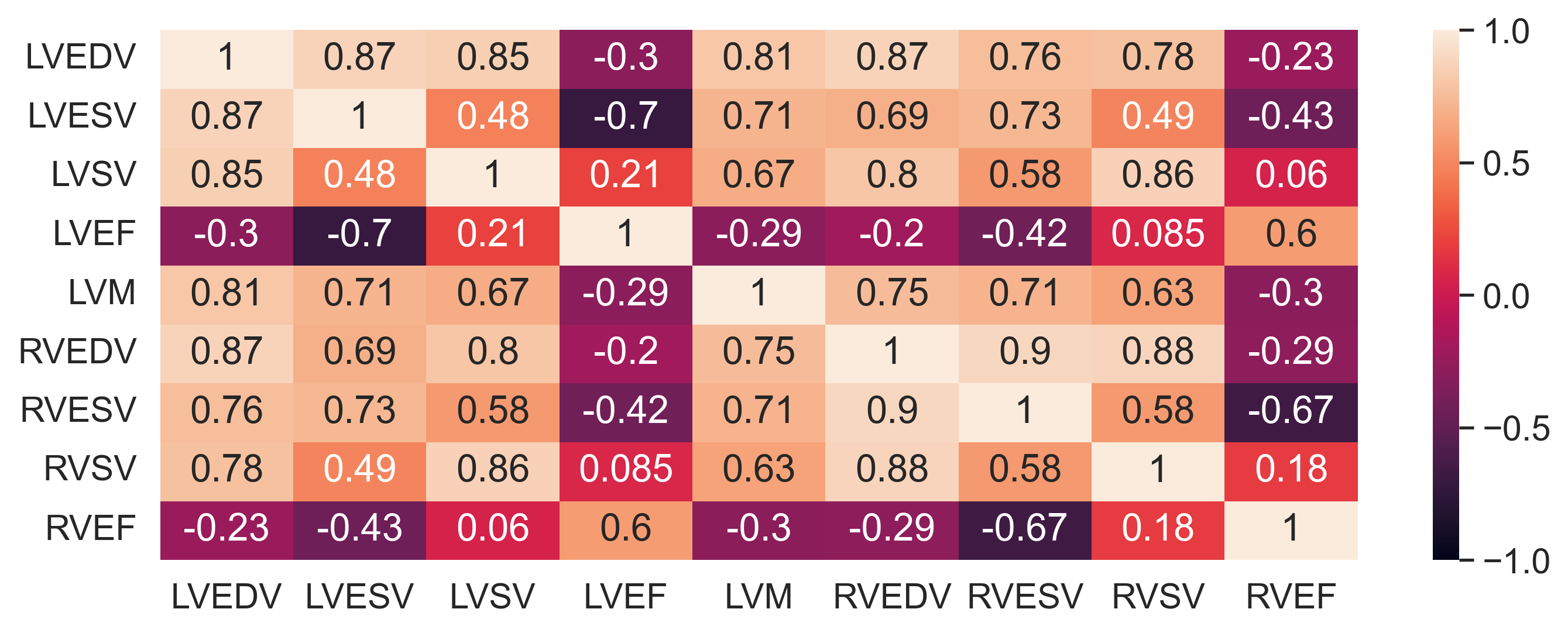}
\caption{Correlation matrix among the used features.}\label{corr}
\end{figure}

The proposed method was implemented in Python 3.9 and it is available online at \url{https://github.com/amaa11/MIP}. To illustrate the degree of collinearity among the used features, Figure~\ref{corr} shows the correlation among the used features. It reveals that most of the used features are correlated with high value either positively or negatively.  Such issue would inevitably affect the outcome of the XAI methods where the collinearity issue is not addressed adequately.
Table~\ref{shap} shows the list of informative predictors generated by the SHAP for each model. It also shows the modified list based on the proposed method. The table shows the order of the features coded into numbers in order to better compare the position of each predictor in each model. The table reveals that the outcome of SHAP is model-dependent as each model generated a different list despite the accuracy of the five models was very close ($\approx$ 0.84). This is due to the fact that each of the used models handles the collinearity in a different way and in consequence SHAP generates different outcomes. This raises a question which one of these lists to consider. However, NMR for each model has different value and the minimum one indicates the more robust list.
\begin{table}[H]
\footnotesize
\centering
\caption{The ordered list of informative predictors generated by SHAP for each used model. The predictors are coded as LVM (1), RVEDV (2), RVESV (3), RVEF (4), LVEDV (5), LVSV (6), LVESV (7), LVEF (8), RVSV (9.}\label{shap}
\begin{tabular}{|ccccc|}
\hline
\multicolumn{5}{|c|}{\textbf{SHAP}}                                                                                                                                                                                                                         \\ \hline
\rowcolor[HTML]{FFCE93} 
\multicolumn{1}{|c|}{\cellcolor[HTML]{FFCE93}\textbf{LBGM}} & \multicolumn{1}{c|}{\cellcolor[HTML]{FFCE93}\textbf{DT}} & \multicolumn{1}{c|}{\cellcolor[HTML]{FFCE93}\textbf{LR}} & \multicolumn{1}{c|}{\cellcolor[HTML]{FFCE93}\textbf{RF}} & \textbf{SVC} \\ \hline
\multicolumn{1}{|c|}{LVM}                                   & \multicolumn{1}{c|}{LVM}                                 & \multicolumn{1}{c|}{LVM}                                 & \multicolumn{1}{c|}{LVM}                                 & LVM          \\ \hline
\multicolumn{1}{|c|}{RVEDV}                                 & \multicolumn{1}{c|}{RVESV}                               & \multicolumn{1}{c|}{RVEDV}                               & \multicolumn{1}{c|}{RVESV}                               & RVEDV        \\ \hline
\multicolumn{1}{|c|}{RVESV}                                 & \multicolumn{1}{c|}{RVEDV}                               & \multicolumn{1}{c|}{RVEF}                                & \multicolumn{1}{c|}{RVEDV}                               & LVEDV        \\ \hline
\multicolumn{1}{|c|}{RVEF}                                  & \multicolumn{1}{c|}{LVSV}                                & \multicolumn{1}{c|}{LVEDV}                               & \multicolumn{1}{c|}{LVEDV}                               & RVEF         \\ \hline
\multicolumn{1}{|c|}{LVEDV}                                 & \multicolumn{1}{c|}{LVESV}                               & \multicolumn{1}{c|}{RVSV}                                & \multicolumn{1}{c|}{LVESV}                               & LVEF         \\ \hline
\multicolumn{1}{|c|}{LVSV}                                  & \multicolumn{1}{c|}{LVEF}                                & \multicolumn{1}{c|}{LVESV}                               & \multicolumn{1}{c|}{RVSV}                                & RVSV         \\ \hline
\multicolumn{1}{|c|}{RVSV}                                  & \multicolumn{1}{c|}{RVSV}                                & \multicolumn{1}{c|}{LVEF}                                & \multicolumn{1}{c|}{RVEF}                                & LVESV        \\ \hline
\multicolumn{1}{|c|}{LVESV}                                 & \multicolumn{1}{c|}{RVEF}                                & \multicolumn{1}{c|}{RVESV}                               & \multicolumn{1}{c|}{LVEF}                                & RVESV        \\ \hline
\multicolumn{1}{|c|}{LVEF}                                  & \multicolumn{1}{c|}{LVEDV}                               & \multicolumn{1}{c|}{LVSV}                                & \multicolumn{1}{c|}{LVSV}                                & LVSV         \\ \hline
\rowcolor[HTML]{FFCE93} 
\multicolumn{1}{|c|}{\cellcolor[HTML]{FFCE93}\textbf{LBGM}} & \multicolumn{1}{c|}{\cellcolor[HTML]{FFCE93}\textbf{DT}} & \multicolumn{1}{c|}{\cellcolor[HTML]{FFCE93}\textbf{LR}} & \multicolumn{1}{c|}{\cellcolor[HTML]{FFCE93}\textbf{RF}} & \textbf{SVC} \\ \hline
\multicolumn{1}{|c|}{1}                                     & \multicolumn{1}{c|}{1}                                   & \multicolumn{1}{c|}{1}                                   & \multicolumn{1}{c|}{1}                                   & 1            \\ \hline
\multicolumn{1}{|c|}{2}                                     & \multicolumn{1}{c|}{3}                                   & \multicolumn{1}{c|}{2}                                   & \multicolumn{1}{c|}{3}                                   & 2            \\ \hline
\multicolumn{1}{|c|}{3}                                     & \multicolumn{1}{c|}{2}                                   & \multicolumn{1}{c|}{4}                                   & \multicolumn{1}{c|}{2}                                   & 5            \\ \hline
\multicolumn{1}{|c|}{4}                                     & \multicolumn{1}{c|}{6}                                   & \multicolumn{1}{c|}{5}                                   & \multicolumn{1}{c|}{5}                                   & 4            \\ \hline
\multicolumn{1}{|c|}{5}                                     & \multicolumn{1}{c|}{7}                                   & \multicolumn{1}{c|}{9}                                   & \multicolumn{1}{c|}{7}                                   & 8            \\ \hline
\multicolumn{1}{|c|}{6}                                     & \multicolumn{1}{c|}{8}                                   & \multicolumn{1}{c|}{7}                                   & \multicolumn{1}{c|}{9}                                   & 9            \\ \hline
\multicolumn{1}{|c|}{9}                                     & \multicolumn{1}{c|}{9}                                   & \multicolumn{1}{c|}{8}                                   & \multicolumn{1}{c|}{4}                                   & 7            \\ \hline
\multicolumn{1}{|c|}{7}                                     & \multicolumn{1}{c|}{4}                                   & \multicolumn{1}{c|}{3}                                   & \multicolumn{1}{c|}{8}                                   & 3            \\ \hline
\multicolumn{1}{|c|}{8}                                     & \multicolumn{1}{c|}{5}                                   & \multicolumn{1}{c|}{6}                                   & \multicolumn{1}{c|}{6}                                   & 6            \\ \hline
\multicolumn{5}{|c|}{\textbf{Modified SHAP}}                                                                                                                                                                                                                \\ \hline
\rowcolor[HTML]{FFCE93} 
\multicolumn{1}{|c|}{\cellcolor[HTML]{FFCE93}\textbf{LBGM}} & \multicolumn{1}{c|}{\cellcolor[HTML]{FFCE93}\textbf{DT}} & \multicolumn{1}{c|}{\cellcolor[HTML]{FFCE93}\textbf{LR}} & \multicolumn{1}{c|}{\cellcolor[HTML]{FFCE93}\textbf{RF}} & \textbf{SVC} \\ \hline
\multicolumn{1}{|c|}{LVM}                                   & \multicolumn{1}{c|}{LVM}                                 & \multicolumn{1}{c|}{LVM}                                 & \multicolumn{1}{c|}{LVM}                                 & LVM          \\ \hline
\multicolumn{1}{|c|}{RVEDV}                                 & \multicolumn{1}{c|}{RVEDV}                               & \multicolumn{1}{c|}{RVEDV}                               & \multicolumn{1}{c|}{RVEDV}                               & RVEDV        \\ \hline
\multicolumn{1}{|c|}{RVESV}                                 & \multicolumn{1}{c|}{RVESV}                               & \multicolumn{1}{c|}{RVSV}                                & \multicolumn{1}{c|}{RVESV}                               & RVSV         \\ \hline
\multicolumn{1}{|c|}{RVSV}                                  & \multicolumn{1}{c|}{LVEDV}                               & \multicolumn{1}{c|}{RVESV}                               & \multicolumn{1}{c|}{LVEDV}                               & RVESV        \\ \hline
\multicolumn{1}{|c|}{LVEDV}                                 & \multicolumn{1}{c|}{LVESV}                               & \multicolumn{1}{c|}{LVEDV}                               & \multicolumn{1}{c|}{LVESV}                               & LVEDV        \\ \hline
\multicolumn{1}{|c|}{LVESV}                                 & \multicolumn{1}{c|}{RVSV}                                & \multicolumn{1}{c|}{RVEF}                                & \multicolumn{1}{c|}{RVSV}                                & RVEF         \\ \hline
\multicolumn{1}{|c|}{RVEF}                                  & \multicolumn{1}{c|}{LVSV}                                & \multicolumn{1}{c|}{LVSV}                                & \multicolumn{1}{c|}{LVSV}                                & LVSV         \\ \hline
\multicolumn{1}{|c|}{LVSV}                                  & \multicolumn{1}{c|}{RVEF}                                & \multicolumn{1}{c|}{LVESV}                               & \multicolumn{1}{c|}{RVEF}                                & LVESV        \\ \hline
\multicolumn{1}{|c|}{LVEF}                                  & \multicolumn{1}{c|}{LVEF}                                & \multicolumn{1}{c|}{LVEF}                                & \multicolumn{1}{c|}{LVEF}                                & LVEF         \\ \hline
\rowcolor[HTML]{FFCE93} 
\multicolumn{1}{|c|}{\cellcolor[HTML]{FFCE93}\textbf{LBGM}} & \multicolumn{1}{c|}{\cellcolor[HTML]{FFCE93}\textbf{DT}} & \multicolumn{1}{c|}{\cellcolor[HTML]{FFCE93}\textbf{LR}} & \multicolumn{1}{c|}{\cellcolor[HTML]{FFCE93}\textbf{RF}} & \textbf{SVC} \\ \hline
\multicolumn{1}{|c|}{1}                                     & \multicolumn{1}{c|}{1}                                   & \multicolumn{1}{c|}{1}                                   & \multicolumn{1}{c|}{1}                                   & 1            \\ \hline
\multicolumn{1}{|c|}{2}                                     & \multicolumn{1}{c|}{2}                                   & \multicolumn{1}{c|}{2}                                   & \multicolumn{1}{c|}{2}                                   & 2            \\ \hline
\multicolumn{1}{|c|}{3}                                     & \multicolumn{1}{c|}{3}                                   & \multicolumn{1}{c|}{9}                                   & \multicolumn{1}{c|}{3}                                   & 9            \\ \hline
\multicolumn{1}{|c|}{9}                                     & \multicolumn{1}{c|}{5}                                   & \multicolumn{1}{c|}{3}                                   & \multicolumn{1}{c|}{5}                                   & 3            \\ \hline
\multicolumn{1}{|c|}{5}                                     & \multicolumn{1}{c|}{7}                                   & \multicolumn{1}{c|}{5}                                   & \multicolumn{1}{c|}{7}                                   & 5            \\ \hline
\multicolumn{1}{|c|}{7}                                     & \multicolumn{1}{c|}{9}                                   & \multicolumn{1}{c|}{4}                                   & \multicolumn{1}{c|}{9}                                   & 4            \\ \hline
\multicolumn{1}{|c|}{4}                                     & \multicolumn{1}{c|}{6}                                   & \multicolumn{1}{c|}{6}                                   & \multicolumn{1}{c|}{6}                                   & 6            \\ \hline
\multicolumn{1}{|c|}{6}                                     & \multicolumn{1}{c|}{4}                                   & \multicolumn{1}{c|}{7}                                   & \multicolumn{1}{c|}{4}                                   & 7            \\ \hline
\multicolumn{1}{|c|}{8}                                     & \multicolumn{1}{c|}{8}                                   & \multicolumn{1}{c|}{8}                                   & \multicolumn{1}{c|}{8}                                   & 8            \\ \hline
\end{tabular}
\end{table}
Table~\ref{example} shows an example of how to reorder the list taking into account the collinearity. The steps start by dividing the index (generated by SHAP) of each predictor by the number of predictors in the model at each iteration, where the top feature is removed. For instance, in the first iteration (j), when there are 9 predictors (NF) in the model, the index (i) of  LVSV is 4 divided by 9, which is the number of predictors leading to a score of 0.44. Then, to calculate the score for the same predictor, we sum up the results of division of the indexes upon the number of features in each iteration as it is explained in equation~\ref{MIP} (0.44+ 0.75+ 0.71+ 0.8+ 1+ 0.5+ 0.33). The scores of the other predictors will be calculated in the same way. The last two columns (orange background) in the table show the score for each predictor in ascending order. The smaller the score, the more the predictor is informative, or, equivalently, the higher the score the less informative the predictor. The score represents how fast (significant) the predictors reached the top of the list in each iteration when the top one was removed. Thereafter, the standard deviation of the scores for each predictor is calculated. This allows identifying the most stable method as the one leading to the largest variance.  The higher SD means that the features are approaching the top list sequentially. It also means that there is a less possibility that group of features are approaching the top list with the similar number of steps (as the case with collinearity). This is also inline with the previously proposed proxy (NMR). The model that produces the lowest NMR value is the more stable one.
\begin{table}[H]
\caption{An example to show how the score of each predictor is calculated. The last row indicates the number of predictors in the model for each iteration.}\label{example}
\begin{tabular}{|cccccccc}
\hline
\multicolumn{8}{|c|}{\textbf{SHAP outcome}}                                                                                                                                                                                                                                                                                                                                                                                                                                                               \\ \hline
\rowcolor[HTML]{EBF1E9} 
\multicolumn{1}{|c|}{\cellcolor[HTML]{EBF1E9}LVM}                                      & \multicolumn{1}{c|}{\cellcolor[HTML]{EBF1E9}0.11}                                      & \multicolumn{1}{c|}{\cellcolor[HTML]{EBF1E9}}      & \multicolumn{1}{c|}{\cellcolor[HTML]{EBF1E9}}     & \multicolumn{1}{c|}{\cellcolor[HTML]{EBF1E9}}      & \multicolumn{1}{c|}{\cellcolor[HTML]{EBF1E9}}     & \multicolumn{1}{c|}{\cellcolor[HTML]{EBF1E9}}      & \multicolumn{1}{c|}{\cellcolor[HTML]{EBF1E9}}    \\ \hline
\rowcolor[HTML]{EBF1E9} 
\multicolumn{1}{|c|}{\cellcolor[HTML]{EBF1E9}RVESV}                                    & \multicolumn{1}{c|}{\cellcolor[HTML]{EBF1E9}0.22}                                      & \multicolumn{1}{c|}{\cellcolor[HTML]{EBF1E9}RVEDV} & \multicolumn{1}{c|}{\cellcolor[HTML]{EBF1E9}0.13} & \multicolumn{1}{c|}{\cellcolor[HTML]{EBF1E9}}      & \multicolumn{1}{c|}{\cellcolor[HTML]{EBF1E9}}     & \multicolumn{1}{c|}{\cellcolor[HTML]{EBF1E9}}      & \multicolumn{1}{c|}{\cellcolor[HTML]{EBF1E9}}    \\ \hline
\rowcolor[HTML]{EBF1E9} 
\multicolumn{1}{|c|}{\cellcolor[HTML]{EBF1E9}RVEDV}                                    & \multicolumn{1}{c|}{\cellcolor[HTML]{EBF1E9}0.33}                                      & \multicolumn{1}{c|}{\cellcolor[HTML]{EBF1E9}RVESV} & \multicolumn{1}{c|}{\cellcolor[HTML]{EBF1E9}0.25} & \multicolumn{1}{c|}{\cellcolor[HTML]{EBF1E9}RVESV} & \multicolumn{1}{c|}{\cellcolor[HTML]{EBF1E9}0.14} & \multicolumn{1}{c|}{\cellcolor[HTML]{EBF1E9}}      & \multicolumn{1}{c|}{\cellcolor[HTML]{EBF1E9}}    \\ \hline
\rowcolor[HTML]{EBF1E9} 
\multicolumn{1}{|c|}{\cellcolor[HTML]{EBF1E9}LVSV}                                     & \multicolumn{1}{c|}{\cellcolor[HTML]{FFCE93}0.44}                                      & \multicolumn{1}{c|}{\cellcolor[HTML]{EBF1E9}LVESV} & \multicolumn{1}{c|}{\cellcolor[HTML]{EBF1E9}0.38} & \multicolumn{1}{c|}{\cellcolor[HTML]{EBF1E9}LVEDV} & \multicolumn{1}{c|}{\cellcolor[HTML]{EBF1E9}0.29} & \multicolumn{1}{c|}{\cellcolor[HTML]{EBF1E9}LVEDV} & \multicolumn{1}{c|}{\cellcolor[HTML]{EBF1E9}0.2} \\ \hline
\rowcolor[HTML]{EBF1E9} 
\multicolumn{1}{|c|}{\cellcolor[HTML]{EBF1E9}LVESV}                                    & \multicolumn{1}{c|}{\cellcolor[HTML]{EBF1E9}0.56}                                      & \multicolumn{1}{c|}{\cellcolor[HTML]{EBF1E9}RVEF}  & \multicolumn{1}{c|}{\cellcolor[HTML]{EBF1E9}0.50} & \multicolumn{1}{c|}{\cellcolor[HTML]{EBF1E9}LVESV} & \multicolumn{1}{c|}{\cellcolor[HTML]{EBF1E9}0.43} & \multicolumn{1}{c|}{\cellcolor[HTML]{EBF1E9}RVEF}  & \multicolumn{1}{c|}{\cellcolor[HTML]{EBF1E9}0.3} \\ \hline
\rowcolor[HTML]{EBF1E9} 
\multicolumn{1}{|c|}{\cellcolor[HTML]{EBF1E9}LVEF}                                     & \multicolumn{1}{c|}{\cellcolor[HTML]{EBF1E9}0.67}                                      & \multicolumn{1}{c|}{\cellcolor[HTML]{EBF1E9}LVEDV} & \multicolumn{1}{c|}{\cellcolor[HTML]{EBF1E9}0.63} & \multicolumn{1}{c|}{\cellcolor[HTML]{EBF1E9}RVEF}  & \multicolumn{1}{c|}{\cellcolor[HTML]{EBF1E9}0.57} & \multicolumn{1}{c|}{\cellcolor[HTML]{EBF1E9}LVESV} & \multicolumn{1}{c|}{\cellcolor[HTML]{EBF1E9}0.5} \\ \hline
\rowcolor[HTML]{EBF1E9} 
\multicolumn{1}{|c|}{\cellcolor[HTML]{EBF1E9}RVSV}                                     & \multicolumn{1}{c|}{\cellcolor[HTML]{EBF1E9}0.78}                                      & \multicolumn{1}{c|}{\cellcolor[HTML]{EBF1E9}LVSV}  & \multicolumn{1}{c|}{\cellcolor[HTML]{FFCE93}0.75} & \multicolumn{1}{c|}{\cellcolor[HTML]{EBF1E9}LVSV}  & \multicolumn{1}{c|}{\cellcolor[HTML]{FFCE93}0.71} & \multicolumn{1}{c|}{\cellcolor[HTML]{EBF1E9}RVSV}  & \multicolumn{1}{c|}{\cellcolor[HTML]{EBF1E9}0.7} \\ \hline
\rowcolor[HTML]{EBF1E9} 
\multicolumn{1}{|c|}{\cellcolor[HTML]{EBF1E9}RVEF}                                     & \multicolumn{1}{c|}{\cellcolor[HTML]{EBF1E9}0.89}                                      & \multicolumn{1}{c|}{\cellcolor[HTML]{EBF1E9}RVSV}  & \multicolumn{1}{c|}{\cellcolor[HTML]{EBF1E9}0.88} & \multicolumn{1}{c|}{\cellcolor[HTML]{EBF1E9}RVSV}  & \multicolumn{1}{c|}{\cellcolor[HTML]{EBF1E9}0.86} & \multicolumn{1}{c|}{\cellcolor[HTML]{EBF1E9}LVSV}  & \multicolumn{1}{c|}{\cellcolor[HTML]{FFCE93}0.8} \\ \hline
\rowcolor[HTML]{EBF1E9} 
\multicolumn{1}{|c|}{\cellcolor[HTML]{EBF1E9}LVEDV}                                    & \multicolumn{1}{c|}{\cellcolor[HTML]{EBF1E9}1.0}                                       & \multicolumn{1}{c|}{\cellcolor[HTML]{EBF1E9}LVEF}  & \multicolumn{1}{c|}{\cellcolor[HTML]{EBF1E9}1.0}  & \multicolumn{1}{c|}{\cellcolor[HTML]{EBF1E9}LVEF}  & \multicolumn{1}{c|}{\cellcolor[HTML]{EBF1E9}1.0}  & \multicolumn{1}{c|}{\cellcolor[HTML]{EBF1E9}LVEF}  & \multicolumn{1}{c|}{\cellcolor[HTML]{EBF1E9}1.0} \\ \hline
\multicolumn{1}{|c|}{9}                                                                & \multicolumn{1}{c|}{}                                                                  & \multicolumn{1}{c|}{8}                             & \multicolumn{1}{c|}{}                             & \multicolumn{1}{c|}{7}                             & \multicolumn{1}{c|}{}                             & \multicolumn{1}{c|}{6}                             & \multicolumn{1}{c|}{}                            \\ \hline
\multicolumn{8}{|c|}{\textbf{SHAP outcome}}                                                                                                                                                                                                                                                                                                                                                                                                                                                               \\ \hline
\rowcolor[HTML]{EBF1E9} 
\multicolumn{1}{|c|}{\cellcolor[HTML]{EBF1E9}RVSV}                                     & \multicolumn{1}{c|}{\cellcolor[HTML]{EBF1E9}0.2}                                       & \multicolumn{1}{c|}{\cellcolor[HTML]{EBF1E9}}      & \multicolumn{1}{c|}{\cellcolor[HTML]{EBF1E9}}     & \multicolumn{1}{c|}{\cellcolor[HTML]{EBF1E9}}      & \multicolumn{1}{c|}{\cellcolor[HTML]{EBF1E9}}     & \multicolumn{1}{c|}{\cellcolor[HTML]{EBF1E9}}      & \multicolumn{1}{c|}{\cellcolor[HTML]{EBF1E9}}    \\ \hline
\rowcolor[HTML]{EBF1E9} 
\multicolumn{1}{|c|}{\cellcolor[HTML]{EBF1E9}LVESV}                                    & \multicolumn{1}{c|}{\cellcolor[HTML]{EBF1E9}0.4}                                       & \multicolumn{1}{c|}{\cellcolor[HTML]{EBF1E9}LVESV} & \multicolumn{1}{c|}{\cellcolor[HTML]{EBF1E9}0.25} & \multicolumn{1}{c|}{\cellcolor[HTML]{EBF1E9}}      & \multicolumn{1}{c|}{\cellcolor[HTML]{EBF1E9}}     & \multicolumn{1}{c|}{\cellcolor[HTML]{EBF1E9}}      & \multicolumn{1}{c|}{\cellcolor[HTML]{EBF1E9}}    \\ \hline
\rowcolor[HTML]{EBF1E9} 
\multicolumn{1}{|c|}{\cellcolor[HTML]{EBF1E9}LVEF}                                     & \multicolumn{1}{c|}{\cellcolor[HTML]{EBF1E9}0.6}                                       & \multicolumn{1}{c|}{\cellcolor[HTML]{EBF1E9}LVSV}  & \multicolumn{1}{c|}{\cellcolor[HTML]{FFCE93}0.5}  & \multicolumn{1}{c|}{\cellcolor[HTML]{EBF1E9}LVSV}  & \multicolumn{1}{c|}{\cellcolor[HTML]{FFCE93}0.33} & \multicolumn{1}{c|}{\cellcolor[HTML]{EBF1E9}}      & \multicolumn{1}{c|}{\cellcolor[HTML]{EBF1E9}}    \\ \hline
\rowcolor[HTML]{EBF1E9} 
\multicolumn{1}{|c|}{\cellcolor[HTML]{EBF1E9}RVEF}                                     & \multicolumn{1}{c|}{\cellcolor[HTML]{EBF1E9}0.8}                                       & \multicolumn{1}{c|}{\cellcolor[HTML]{EBF1E9}LVEF}  & \multicolumn{1}{c|}{\cellcolor[HTML]{EBF1E9}0.75} & \multicolumn{1}{c|}{\cellcolor[HTML]{EBF1E9}LVEF}  & \multicolumn{1}{c|}{\cellcolor[HTML]{EBF1E9}0.67} & \multicolumn{1}{c|}{\cellcolor[HTML]{EBF1E9}RVEF}  & \multicolumn{1}{c|}{\cellcolor[HTML]{EBF1E9}0.5} \\ \hline
\rowcolor[HTML]{EBF1E9} 
\multicolumn{1}{|c|}{\cellcolor[HTML]{EBF1E9}LVSV}                                     & \multicolumn{1}{c|}{\cellcolor[HTML]{FFCE93}1}                                         & \multicolumn{1}{c|}{\cellcolor[HTML]{EBF1E9}RVEF}  & \multicolumn{1}{c|}{\cellcolor[HTML]{EBF1E9}1}    & \multicolumn{1}{c|}{\cellcolor[HTML]{EBF1E9}RVEF}  & \multicolumn{1}{c|}{\cellcolor[HTML]{EBF1E9}1.00} & \multicolumn{1}{c|}{\cellcolor[HTML]{EBF1E9}LVEF}  & \multicolumn{1}{c|}{\cellcolor[HTML]{EBF1E9}1}   \\ \hline
\multicolumn{1}{|c|}{5}                                                                & \multicolumn{1}{c|}{}                                                                  & \multicolumn{1}{c|}{4}                             & \multicolumn{1}{c|}{}                             & \multicolumn{1}{c|}{3}                             & \multicolumn{1}{c|}{}                             & \multicolumn{1}{c|}{2}                             & \multicolumn{1}{c|}{}                            \\ \hline
\multicolumn{3}{|c|}{\textbf{}}                                                                                                                                                                                                      &                                                   &                                                    &                                                   &                                                    &                                                  \\ \cline{1-3}
\multicolumn{1}{|c|}{\textbf{\begin{tabular}[c]{@{}c@{}}SHAP \\ outcome\end{tabular}}} & \multicolumn{1}{c|}{\textbf{\begin{tabular}[c]{@{}c@{}}Modified \\ SHAP\end{tabular}}} & \multicolumn{1}{c|}{\textbf{Score}}                & \multicolumn{1}{l}{}                              & \multicolumn{1}{l}{}                               & \multicolumn{1}{l}{}                              & \multicolumn{1}{l}{}                               & \multicolumn{1}{l}{}                             \\ \cline{1-3}
\multicolumn{1}{|c|}{\cellcolor[HTML]{EBF1E9}LVM}                                      & \multicolumn{1}{c|}{\cellcolor[HTML]{FFCE93}LVM}                                       & \multicolumn{1}{c|}{0.11}                          & \multicolumn{1}{l}{}                              & \multicolumn{1}{l}{}                               & \multicolumn{1}{l}{}                              & \multicolumn{1}{l}{}                               & \multicolumn{1}{l}{}                             \\ \cline{1-3}
\multicolumn{1}{|c|}{\cellcolor[HTML]{EBF1E9}RVESV}                                    & \multicolumn{1}{c|}{\cellcolor[HTML]{FFCE93}RVEDV}                                     & \multicolumn{1}{c|}{0.46}                          & \multicolumn{1}{l}{}                              & \multicolumn{1}{l}{}                               & \multicolumn{1}{l}{}                              & \multicolumn{1}{l}{}                               & \multicolumn{1}{l}{}                             \\ \cline{1-3}
\multicolumn{1}{|c|}{\cellcolor[HTML]{EBF1E9}RVEDV}                                    & \multicolumn{1}{c|}{\cellcolor[HTML]{FFCE93}RVESV}                                     & \multicolumn{1}{c|}{0.47}                          & \multicolumn{1}{l}{}                              & \multicolumn{1}{l}{}                               & \multicolumn{1}{l}{}                              & \multicolumn{1}{l}{}                               & \multicolumn{1}{l}{}                             \\ \cline{1-3}
\multicolumn{1}{|c|}{\cellcolor[HTML]{EBF1E9}LVSV}                                     & \multicolumn{1}{c|}{\cellcolor[HTML]{FFCE93}LVEDV}                                     & \multicolumn{1}{c|}{2.12}                          & \multicolumn{1}{l}{}                              & \multicolumn{1}{l}{}                               & \multicolumn{1}{l}{}                              & \multicolumn{1}{l}{}                               & \multicolumn{1}{l}{}                             \\ \cline{1-3}
\multicolumn{1}{|c|}{\cellcolor[HTML]{EBF1E9}LVESV}                                    & \multicolumn{1}{c|}{\cellcolor[HTML]{FFCE93}LVESV}                                     & \multicolumn{1}{c|}{2.52}                          & \multicolumn{1}{l}{}                              & \multicolumn{1}{l}{}                               & \multicolumn{1}{l}{}                              & \multicolumn{1}{l}{}                               & \multicolumn{1}{l}{}                             \\ \cline{1-3}
\multicolumn{1}{|c|}{\cellcolor[HTML]{EBF1E9}LVEF}                                     & \multicolumn{1}{c|}{\cellcolor[HTML]{FFCE93}RVSV}                                      & \multicolumn{1}{c|}{3.42}                          & \multicolumn{1}{l}{}                              & \multicolumn{1}{l}{}                               & \multicolumn{1}{l}{}                              & \multicolumn{1}{l}{}                               & \multicolumn{1}{l}{}                             \\ \cline{1-3}
\multicolumn{1}{|c|}{\cellcolor[HTML]{EBF1E9}RVSV}                                     & \multicolumn{1}{c|}{\cellcolor[HTML]{FFCE93}LVSV}                                      & \multicolumn{1}{c|}{4.53}                          & \multicolumn{1}{l}{}                              & \multicolumn{1}{l}{}                               & \multicolumn{1}{l}{}                              & \multicolumn{1}{l}{}                               & \multicolumn{1}{l}{}                             \\ \cline{1-3}
\multicolumn{1}{|c|}{\cellcolor[HTML]{EBF1E9}RVEF}                                     & \multicolumn{1}{c|}{\cellcolor[HTML]{FFCE93}RVEF}                                      & \multicolumn{1}{c|}{5.56}                          & \multicolumn{1}{l}{}                              & \multicolumn{1}{l}{}                               & \multicolumn{1}{l}{}                              & \multicolumn{1}{l}{}                               & \multicolumn{1}{l}{}                             \\ \cline{1-3}
\multicolumn{1}{|c|}{\cellcolor[HTML]{EBF1E9}LVEDV}                                    & \multicolumn{1}{c|}{\cellcolor[HTML]{FFCE93}LVEF}                                      & \multicolumn{1}{c|}{6.69}                          & \multicolumn{1}{l}{}                              & \multicolumn{1}{l}{}                               & \multicolumn{1}{l}{}                              & \multicolumn{1}{l}{}                               & \multicolumn{1}{l}{}                             \\ \cline{1-3}
\end{tabular}
\end{table}
Table~\ref{NMRSD} shows the SD of the scores of predictors for each model. The maximum SD value is 2.71 for RF model. This indicates that the modified SHAP list is more stable for the RF model compared to other models. This is also inline with the NMR value for each model, reaching the minimum for the RF model. 
\begin{table}[H]
\centering
\caption{The NMR and the standard deviation (SD) for each model.}\label{NMRSD}
\begin{tabular}{|cc|cc|}
\hline
\multicolumn{2}{|c|}{\cellcolor[HTML]{CBCEFB}\textbf{SD}}                                                & \multicolumn{2}{c|}{\cellcolor[HTML]{CBCEFB}\textbf{NMR}}                   \\ \hline
\multicolumn{1}{|c|}{\textbf{Model}} & \textbf{\begin{tabular}[c]{@{}c@{}}Modified \\ SHAP\end{tabular}} & \multicolumn{1}{c|}{\textbf{Model}} & \textbf{SHAP}                         \\ \hline
\multicolumn{1}{|c|}{DT}             & 2.36                                                              & \multicolumn{1}{c|}{DT}             & 0.43                                  \\ \hline
\multicolumn{1}{|c|}{LBGM}           & 2.41                                                              & \multicolumn{1}{c|}{LBGM}           & 0.49                                  \\ \hline
\multicolumn{1}{|c|}{LR}             & 2.11                                                              & \multicolumn{1}{c|}{LR}             & 0.57                                  \\ \hline
\multicolumn{1}{|c|}{RF}             & \cellcolor[HTML]{FFE699}\textbf{2.71}                             & \multicolumn{1}{c|}{RF}             & \cellcolor[HTML]{FFE699}\textbf{0.14} \\ \hline
\multicolumn{1}{|c|}{SVC}            & 2.16                                                              & \multicolumn{1}{c|}{SVC}            & 0.58                                  \\ \hline
\end{tabular}
\end{table}
To reveal the impact of dependency among the predictors, Table~\ref{PCA} shows the ordered list the components generated by SHAP for the five used models. It shows that four models (RF, LBGM, SVC and LR) share the exact same outcome, while DT is slightly different. In addition, the table shows that the outcome of the proposed method is exactly the same of the SHAP outcome in the four models, while it is slightly miss-matched for the DT model. The results confirm the hypothesis that when the predictors are independent the outcome of the proposed method will be similar to the outcome of SHAP.
\begin{table}[H]
\centering
\caption{The order list of informative predictors generated by SHAP for each used model using the PCs.}\label{PCA}
\begin{tabular}{|cc|cc|}
\hline
\multicolumn{2}{|c|}{\textbf{DT}}                       & \multicolumn{2}{c|}{\textbf{RF, LBGM, SVC and LR}}     \\ \hline
\multicolumn{1}{|c|}{\textbf{Proposed}} & \textbf{SHAP} & \multicolumn{1}{c|}{\textbf{Proposed}} & \textbf{SHAP} \\ \hline
\multicolumn{1}{|c|}{PC1}               & PC1           & \multicolumn{1}{c|}{PC1}               & PC1           \\ \hline
\multicolumn{1}{|c|}{PC4}               & PC4           & \multicolumn{1}{c|}{PC4}               & PC4           \\ \hline
\multicolumn{1}{|c|}{PC2}               & PC3           & \multicolumn{1}{c|}{PC2}               & PC2           \\ \hline
\multicolumn{1}{|c|}{PC3}               & PC2           & \multicolumn{1}{c|}{PC3}               & PC3           \\ \hline
\end{tabular}
\end{table}
The orders of the nine features with the codes explained in Table~\ref{benchmark}. The Kendall's tau\_b rank correlation value between the benchmark list and SHAP was 0.38 (p-value=0.1) while the correlation between the benchmark list and the proposed method was 0.61 (p-value = 0.02). In addition, the Pearson's Correlation between the benchmark list and SHAP list was 0.58 (p-value = 0.09) while the correlation value between the benchmark order and the proposed method was 0.70 (p-value= 0.03).
\begin{table}[H]
\centering
\caption{The ordered list of informative predictors based based the benchmark, SHAP and the proposed method. The predictors are coded as LVM (1), RVESV (2), LVESV (3), LVEDV (4), LVSV (5), RVEDV (6), RVSV (7), RVEF (8), LVEF (9).}\label{benchmark}
\begin{tabular}{|ccc|ccc|}
\hline
\multicolumn{3}{|c|}{\textbf{Actual order}}                                                      & \multicolumn{3}{c|}{\textbf{Coded order}}                                                       \\ \hline
\multicolumn{1}{|c|}{\textbf{benchmark}} & \multicolumn{1}{c|}{\textbf{SHAP}} & \textbf{Proposed} & \multicolumn{1}{c|}{\textbf{benchmark}} & \multicolumn{1}{c|}{\textbf{SHAP}} & \textbf{Proposed} \\ \hline
\multicolumn{1}{|c|}{RVESV}             & \multicolumn{1}{c|}{LVM}           & LVM               & \multicolumn{1}{c|}{2}                 & \multicolumn{1}{c|}{1}             & 1                 \\ \hline
\multicolumn{1}{|c|}{LVM}               & \multicolumn{1}{c|}{RVESV}         & RVEDV             & \multicolumn{1}{c|}{1}                 & \multicolumn{1}{c|}{2}             & 6                 \\ \hline
\multicolumn{1}{|c|}{LVESV}             & \multicolumn{1}{c|}{RVEDV}         & RVESV             & \multicolumn{1}{c|}{3}                 & \multicolumn{1}{c|}{6}             & 2                 \\ \hline
\multicolumn{1}{|c|}{LVEDV}             & \multicolumn{1}{c|}{LVSV}          & LVEDV             & \multicolumn{1}{c|}{4}                 & \multicolumn{1}{c|}{5}             & 4                 \\ \hline
\multicolumn{1}{|c|}{LVSV}              & \multicolumn{1}{c|}{LVESV}         & LVESV             & \multicolumn{1}{c|}{5}                 & \multicolumn{1}{c|}{3}             & 3                 \\ \hline
\multicolumn{1}{|c|}{RVEDV}             & \multicolumn{1}{c|}{LVEF}          & RVSV              & \multicolumn{1}{c|}{6}                 & \multicolumn{1}{c|}{9}             & 7                 \\ \hline
\multicolumn{1}{|c|}{RVSV}              & \multicolumn{1}{c|}{RVSV}          & LVSV              & \multicolumn{1}{c|}{7}                 & \multicolumn{1}{c|}{7}             & 5                 \\ \hline
\multicolumn{1}{|c|}{RVEF}              & \multicolumn{1}{c|}{RVEF}          & RVEF              & \multicolumn{1}{c|}{8}                 & \multicolumn{1}{c|}{8}             & 8                 \\ \hline
\multicolumn{1}{|c|}{LVEF}              & \multicolumn{1}{c|}{LVEDV}         & LVEF              & \multicolumn{1}{c|}{9}                 & \multicolumn{1}{c|}{4}             & 9                 \\ \hline
\end{tabular}
\end{table}
\section{Discussions}
In real life applications the predictors are prone to a high degree of collinearity. Such issue must be considered and appropriately addressed specially in medical and biological domains, where miss-identifying the informative variables might lead to serious consequences. Another relevant aspect of XAI methods that should be considered is the simplicity which facilitates understandability. Most of the end-users are not experts in the machine learning models neither in the XAI approaches, they are rather lay-users. Accordingly, it is vital to present methods and approaches to explain the model simply in order for the users to understand and implement them correctly. We presented a simple method that modifies the XAI outcomes to consider the dependency among the predictors and lead to a more robust list. The proposed approach is based on the fact that if two predictors in the model are collinear and they both affect the outcome to a similar degree, then they should get similar significance in the list of informative predictors. This objective is achieved by removing the top predictor in the model, re-training and testing the model, and checking the positions of the predictors. The proposed method is model-agnostic, which means that it can be applied to any XAI method that does not consider the collinearity among the predictors in its outcome.\\
Comparing the proposed method with what have been proposed before, ours has several advantages over other methods. Firstly, the proposed method is simple, easy to catch and understand as it does not require to generate new distributions or cohorts to remove the effect of the dependency as it is the case in~\cite{aas2021explaining} and~\cite{mase2019explaining}. Secondly, in~\cite{aas2021explaining} the chosen distribution might affect the SHAP outcome. Similarly, in~\cite{mase2019explaining} the chosen similarity matrics will affect the new generated cohort. In other words, the two methods are user-dependent requiring to select the distribution or the similarity matrics that fits the data. However, the end-user may not be familiar with such issue and might chose the wrong option which will affect the quality of the outcome. The proposed method, it is not user-dependent and the user has nothing to do with the quality of the outcome. Thirdly, the two proposed methods are providing local explanation, meaning they can explain the model for one single subject or group of subjects and the outcome cannot be interpreted for the whole instances. The proposed method can be applied locally and globally as long as there is a list of informative predictors. In the local explanation, the user can follow the same steps illustrated in the ~\ref{example} to generate new list of informative predictors taken into account the collinearity. 
In the proposed method, we followed functionality-grounded evaluation. The proposed proxy as such evaluation is neither biased nor expensive as it is the case for application-grounded evaluation~\cite{zhou2021evaluating}.\\ 
In terms of clinical validation, the results of the correlation analysis indicate that the order of the features based on the proposed method is more inline with the reference (benchmark) order derived from biological priors than the SHAP list. Moreover, Zahra Raisi-Estabragh et. al.,~\cite{raisi2021variation} published the mean difference between men and women of four of the used features that are LVM, LVESV, LVEDV and LVEF. Their results indicate that the mean difference in LVM is the higher followed by LVEDV and LVESV while LVEF is with a very small difference which made it to occupy the bottom of the list. The results are matching with the proposed method in a way that the order is exactly the same. This confirms that the proposed methods is more in line with the benchmark side than SHAP.\\
The main limitation of the proposed method is that it is XAI model-dependent. MIP cannot identify the top features, it rather depends on the XAI method (e.g., SHAP) to highlight them, then it modifies the order to consider the collinearity. Secondly, if the used predictors are independent, the outcome will not be different from other XAI methods (or the classifiers). This point could be considered either a weak or a strength side of the proposed method. To consider it as a weak point, if the predictors are independent, then the user might abandon to implement it. On the other hand, it might be better to consider it as a confirmation and to compare it with the outcome of XAI method even if the predictors are independent. 

\section{Conclusions}
To conclude, we presented a simple, yet useful measure that take into account the collinearity when XAI methods applied to any machine learning model. The presented measure will help the users to better use XAI methods and produce more meaningful explanation. It can be applied to any XAI method that does not take into account the collinearity among the predictors. It can be used either locally for a single subject or globally to show the marginal contribution of the features for all instances considering the collinearity. The proposed method deals with one of the main issue in real life application to handle it and presents more robust results.
\section{Funding}
AS is supported by a British Heart Foundation project grant (PG/21/10619). IBG and GM acknowledge support from Fondazione CariVerona (Bando Ricerca Scientifica di Eccellenza 2018, EDIPO project - reference number 2018.0855.2019. ZRE recognises the National Institute for Health and Care Research (NIHR) Integrated Academic Training programme which supports her Academic Clinical Lectureship post and was also supported by British Heart Foundation Clinical Research Training Fellowship No. FS/17/81 /33318. SEP acknowledges support from the National Institute for Health and Care Research (NIHR) Biomedical Research Centre at Barts and have received funding from the European Union’s Horizon 2020 research and innovation programme under grant agreement No 825903 (euCanSHare project). This article is supported by the London Medical Imaging and Artificial Intelligence Centre for Value Based Healthcare (AI4VBH), which is funded from the Data to Early Diagnosis and Precision Medicine strand of the government’s Industrial Strategy Challenge Fund, managed and delivered by Innovate UK on behalf of UK Research and Innovation (UKRI). Views expressed are those of the authors and not necessarily those of the AI4VBH Consortium members, the NHS, Innovate UK, or UKRI.
\section{Disclosure}
SEP provides consultancy to Cardiovascular Imaging Inc, Calgary, Alberta, Canada. The remaining authors have no disclosures.

\label{}



\bibliographystyle{unsrt} 
\bibliography{ref.bib}

\begin{thebibliography}{10}

\bibitem{gunning2019xai}
David Gunning, Mark Stefik, Jaesik Choi, Timothy Miller, Simone Stumpf, and
  Guang-Zhong Yang.
\newblock Xai—explainable artificial intelligence.
\newblock {\em Science robotics}, 4(37):eaay7120, 2019.

\bibitem{szabo2022clinician}
Liliana Szabo, Zahra Raisi-Estabragh, Ahmed Salih, Celeste McCracken, Esmeralda
  Ruiz~Pujadas, Polyxeni Gkontra, Mate Kiss, Pal Maurovich-Horvath, Hajnalka
  Vago, Bela Merkely, et~al.
\newblock Clinician's guide to trustworthy and responsible artificial
  intelligence in cardiovascular imaging.
\newblock {\em Frontiers in Cardiovascular Medicine, 2022, vol. 9}, 2022.

\bibitem{lundberg2017unified}
Scott~M Lundberg and Su-In Lee.
\newblock A unified approach to interpreting model predictions.
\newblock {\em Advances in neural information processing systems}, 30, 2017.

\bibitem{ribeiro2016should}
Marco~Tulio Ribeiro, Sameer Singh, and Carlos Guestrin.
\newblock " why should i trust you?" explaining the predictions of any
  classifier.
\newblock In {\em Proceedings of the 22nd ACM SIGKDD international conference
  on knowledge discovery and data mining}, pages 1135--1144, 2016.

\bibitem{friedman2001greedy}
Jerome~H Friedman.
\newblock Greedy function approximation: a gradient boosting machine.
\newblock {\em Annals of statistics}, pages 1189--1232, 2001.

\bibitem{belle2021principles}
Vaishak Belle and Ioannis Papantonis.
\newblock Principles and practice of explainable machine learning.
\newblock {\em Frontiers in big Data}, page~39, 2021.

\bibitem{aas2021explaining}
Kjersti Aas, Martin Jullum, and Anders L{\o}land.
\newblock Explaining individual predictions when features are dependent: More
  accurate approximations to shapley values.
\newblock {\em Artificial Intelligence}, 298:103502, 2021.

\bibitem{mase2019explaining}
Masayoshi Mase, Art~B Owen, and Benjamin Seiler.
\newblock Explaining black box decisions by shapley cohort refinement.
\newblock {\em arXiv preprint arXiv:1911.00467}, 2019.

\bibitem{salih2022investigating}
Ahmed Salih, Ilaria~Boscolo Galazzo, Federica Cruciani, Lorenza Brusini, and
  Petia Radeva.
\newblock Investigating explainable artificial intelligence for mri-based
  classification of dementia: a new stability criterion for explainable
  methods.
\newblock In {\em 2022 IEEE International Conference on Image Processing
  (ICIP)}, pages 4003--4007. IEEE, 2022.

\bibitem{petersen2017reference}
Steffen~E Petersen, Nay Aung, Mihir~M Sanghvi, Filip Zemrak, Kenneth Fung,
  Jose~Miguel Paiva, Jane~M Francis, Mohammed~Y Khanji, Elena Lukaschuk,
  Aaron~M Lee, et~al.
\newblock Reference ranges for cardiac structure and function using
  cardiovascular magnetic resonance (cmr) in caucasians from the uk biobank
  population cohort.
\newblock {\em Journal of Cardiovascular Magnetic Resonance}, 19(1):1--19,
  2017.

\bibitem{st2022sex}
Sarah~R St~Pierre, Mathias Peirlinck, and Ellen Kuhl.
\newblock Sex matters: a comprehensive comparison of female and male hearts.
\newblock {\em Frontiers in Physiology}, 13:303, 2022.

\bibitem{zhou2021evaluating}
Jianlong Zhou, Amir~H Gandomi, Fang Chen, and Andreas Holzinger.
\newblock Evaluating the quality of machine learning explanations: A survey on
  methods and metrics.
\newblock {\em Electronics}, 10(5):593, 2021.

\bibitem{raisi2021variation}
Zahra Raisi-Estabragh, Asmaa~AM Kenawy, Nay Aung, Jackie Cooper, Patricia~B
  Munroe, Nicholas~C Harvey, Steffen~E Petersen, and Mohammed~Y Khanji.
\newblock Variation in left ventricular cardiac magnetic resonance normal
  reference ranges: systematic review and meta-analysis.
\newblock {\em European Heart Journal-Cardiovascular Imaging}, 22(5):494--504,
  2021.

\end{thebibliography}




\end{document}